\documentclass[lettersize,journal, twoside]{IEEEtran}
\usepackage{amsmath,amsfonts}
\usepackage{algorithmic}
\usepackage{algorithm}
\usepackage{array}
\usepackage[caption=false,font=normalsize,labelfont=sf,textfont=sf]{subfig}
\usepackage{textcomp}
\usepackage{stfloats}
\usepackage{url}
\usepackage{verbatim}
\usepackage{graphicx}

\usepackage{amsmath}
\usepackage{amsfonts}
\usepackage{multirow}
\usepackage{graphicx}
\usepackage{booktabs}
\usepackage{wrapfig}
\usepackage{tabularray}
\usepackage{xcolor}
\usepackage{subcaption}
\usepackage[table]{xcolor}

\usepackage{pifont}

\usepackage{algorithmic}
\usepackage{algorithm}
\usepackage{array}
\usepackage[caption=false,font=normalsize,labelfont=sf,textfont=sf]{subfig}
\usepackage{textcomp}
\usepackage{stfloats}
\usepackage{url}
\usepackage{verbatim}
\definecolor{darkgreen}{RGB}{0, 154, 85}
\usepackage[compress]{cite}
\usepackage[capitalize]{cleveref}

\begin{document}

\title{\LARGE \bf
Risk Map As Middleware: Towards Interpretable Cooperative  \\ End-to-end Autonomous Driving for Risk-Aware Planning}

\author{Mingyue Lei$^{1,2*}$, Zewei Zhou$^{2*}$, Hongchen Li$^{1}$, Jiaqi Ma$^{2\dagger}$ and Jia Hu$^{1\dagger}$
\thanks{Manuscript received: July 5, 2025; Revised: September 1, 2025; Accepted: October 31, 2025. This paper was recommended for publication by Editor Pascal Vasseur upon evaluation of the Associate Editor and Reviewers’ comments. This paper is partially supported by USDOT/FHWA Mobility Center of Excellence, National Science Foundation \# 2346267 POSE: Phase II: DriveX, National Science and Technology Major Project (No. 2022ZD0115501 or 2022ZD0115503), National Key R\&D Program of China (2022YFF0604905 or 2022YFE0117100), National Natural Science Foundation of China (Grant No. 52372317), Yangtze River Delta Science and Technology Innovation Joint Force (No. 2023CSJGG0800), Shanghai Automotive Industry Science and Technology Development Foundation (No. 2404), Xiaomi Young Talents Program, the Fundamental Research Funds for the Central Universities (22120230311), and Tongji Zhongte Chair Professor Foundation (No. 000000375-2018082). $^{*}$Equally contributed to the work. $^{\dagger}$Corresponding author.}
\thanks{$^{1}$Mingyue Lei, Hongchen Li and Jia Hu are with the Key Laboratory of Road and Traffic Engineering of the Ministry of Education, Tongji University, Shanghai, China. 
        {\tt\small mingyue\_l@tongji.edu.cn, 2410195@tongji.edu.cn, hujia@tongji.edu.cn}}%
\thanks{$^{2}$Mingyue Lei, Zewei Zhou and Jiaqi Ma are with UCLA Mobility Lab, University of California, Los Angeles, USA. This work was done during Mingyue Lei's visit at UCLA Mobility Lab.
        {\tt\small mingyue2024@ucla.edu, zeweizhou@ucla.edu, jiaqima@ucla.edu}}%
\thanks{Digital Object Identifier (DOI): see top of this page.}
}

\markboth{IEEE ROBOTICS AND AUTOMATION LETTERS. PREPRINT VERSION. NOVEMBER, 2025}%
{Lei \MakeLowercase{\textit{et al.}}: Risk Map As Middleware: Towards Interpretable Cooperative End-to-end Autonomous Driving for Risk-Aware Planning}


\maketitle

\begin{abstract}
  End-to-end paradigm has emerged as a promising approach to autonomous driving. However, existing single-agent end-to-end pipelines are often constrained by occlusion and limited perception range, resulting in hazardous driving. Furthermore, their black-box nature prevents the interpretability of the driving behavior, leading to an untrustworthiness system. 
  To address these limitations, we introduce \textit{Risk Map as Middleware (RiskMM)} and propose an interpretable cooperative end-to-end driving framework. 
  The risk map learns directly from the driving data and provides an interpretable spatiotemporal representation of the scenario from the upstream perception and the interactions between the ego vehicle and the surrounding environment for downstream planning.
  RiskMM first constructs a multi-agent spatiotemporal representation with unified Transformer-based architecture, then derives risk-aware representations by modeling interactions among surrounding environments with attention. These representations are subsequently fed into a learning-based Model Predictive Control (MPC) module. The MPC planner inherently accommodates physical constraints and different vehicle types and can provide interpretation by aligning learned parameters with explicit MPC elements. Evaluations conducted on the real-world V2XPnP-Seq dataset confirm that RiskMM achieves superior and robust performance in risk-aware trajectory planning, significantly enhancing the interpretability of the cooperative end-to-end driving framework. The codebase will be released to facilitate future research in this field.
\end{abstract}


\section{INTRODUCTION}
\label{sec:intro}

\begin{figure}[t]
  \centering
  \includegraphics[width=\linewidth]{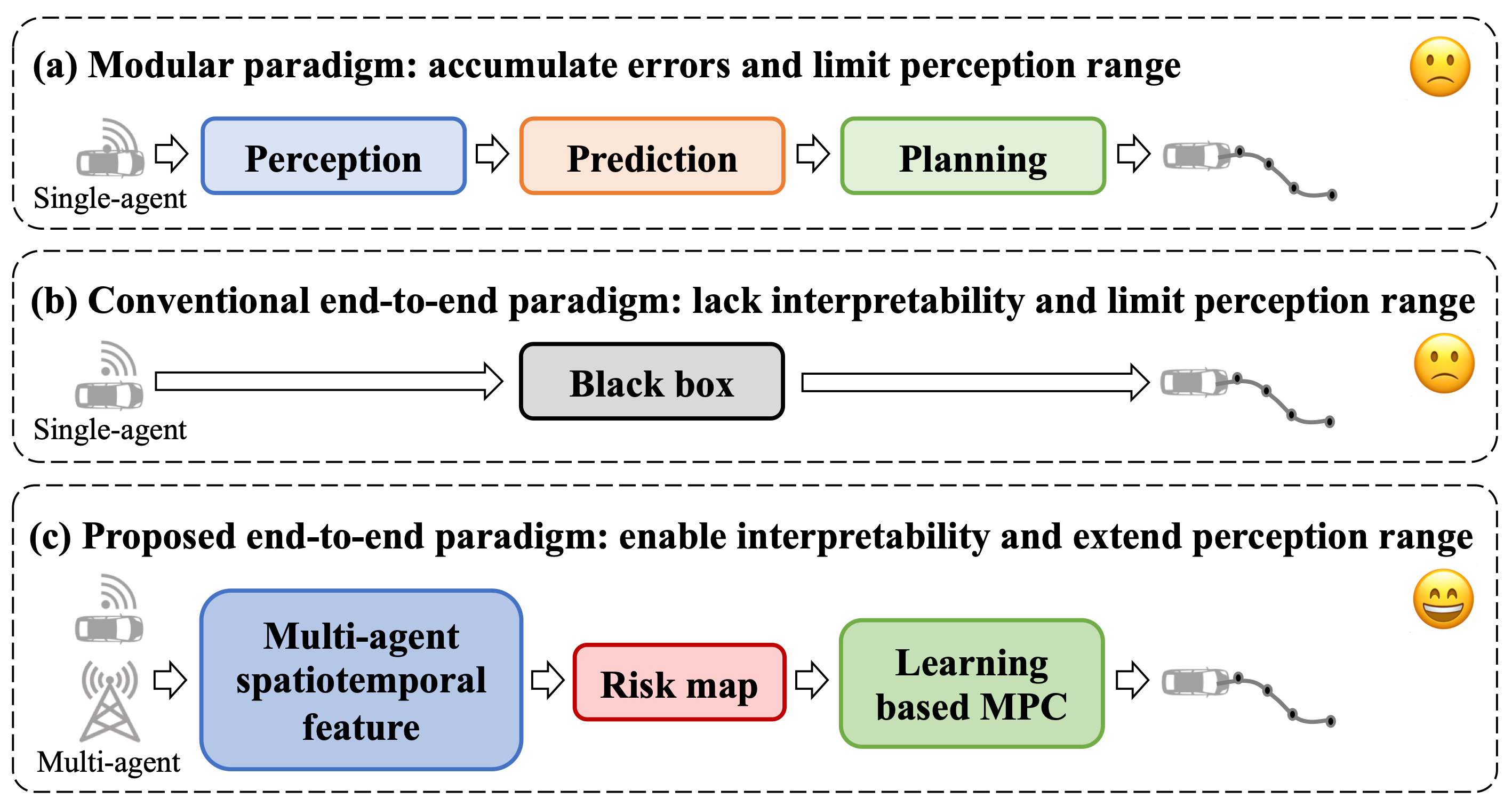}
  \caption{Comparison of different autonomous driving paradigms. \textit{(a) Modular paradigm}, which suffers from error propagation and the limited perception of single-vehicle systems. \textit{(b) Conventional end-to-end paradigm}, which directly maps the single-vehicle sensor input to control commands but lacks interpretability and holistic perception. \textit{(c) Proposed end-to-end paradigm}, which introduces the risk map as the middleware and integrates the multi-agent cooperation to enhance interpretability and overcome perception constraints.} 
  \label{Figure1}
  \vspace{-0.7cm}
\end{figure}

As a safety-critical system, the autonomous driving system has evolved from a modular paradigm with distinct perception, prediction, and planning modules to an end-to-end framework \cite{hu2023planning, zheng2024genad}, eliminating error accumulation in the modular design and enabling joint optimization of the final planning performance. Nevertheless, two major challenges persist: \textit{1) Occlusion and limited perception range inherent to single-vehicle systems}, and \textit{2) Black-box end-to-end framework that hinders the generation of safe and interpretable planning trajectories}.


First, existing end-to-end autonomous driving paradigms mainly focus on single-vehicle systems \cite{liao2024diffusiondrive,weng2024drive}. Although error propagation between separate modules is mitigated by the end-to-end framework, these systems remain constrained by the limited perceptual range and perception noise in input data \cite{chen2024end}. Thus, multi-agent systems have gained prominence, which enable connected and automated vehicles (CAVs) and infrastructure units to share complementary information with Vehicle-to-Everything (V2X) communication. However, how to capture the spatiotemporal scenario representation and how to adopt it for diverse tasks become more intricate when scaling to multiple agents. The modular paradigm remains the popular approach in cooperative driving \cite{lei2025cooperrisk, wang2025cmp}. While partial multi-agent end-to-end models have been investigated, including joint detection-prediction \cite{zhou2024v2xpnp} or detection-tracking frameworks \cite{xu2025learning}, a comprehensive multi-agent end-to-end framework for final planning still requires exploration.


Moreover, trajectory generation, a core component of end-to-end autonomous driving, determines the vehicle's final physical actions and must rigorously incorporate physical constraints such as vehicle dynamics and safety considerations \cite{bairouk2024exploring}. Furthermore, for the safety-critical autonomous driving system, the trajectory generation must maintain interpretability and transparency to ensure safety and to support focused improvements on specific components and performance \cite{chitta2021neat, chen2021interpretable}. Current research mainly focuses on enhancing planning accuracy while overlooking the interpretability of generated trajectories \cite{hu2023planning,liao2024diffusiondrive}.

To address these challenges, we propose an interpretable cooperative end-to-end
driving framework by incorporating a \textbf{\textit{Risk Map as Middleware} (RiskMM)}. The proposed RiskMM explicitly captures the spatial-temporal distribution of the driving risk surrounding the ego vehicle. Furthermore, we integrate a learning-based Model Predictive Control (MPC) module that utilizes the risk map to generate planning trajectories under explicit driving constraints, enhancing interpretability. Specifically, our method begins by extracting multi-agent spatiotemporal scenario representations through auxiliary cooperative detection and prediction tasks. Subsequently, the RiskMM and trajectory planning modules directly learn from the scenario representations with the supervision signal of planning trajectories, facilitating end-to-end optimization and yielding the corresponding explanation.

The risk map middleware explicitly models interaction-induced risk distributions between the ego vehicle and the surroundings based on the spatiotemporal feature of scenario representation. It not only enhances interpretability but also supports scenario assessment, which is conventionally dependent on rule-based risk quantification. Notably, the risk map is learned and optimized in an end-to-end paradigm, rather than being manually crafted \cite{lei2025cooperrisk}. Moreover, the parameters learned within the MPC module inherently represent the driving styles, such as aggressive or conservative ones, enabling strategy adaptation to various scenarios. Additionally, our MPC design can explicitly integrate driving constraints and accommodate varying vehicle dynamics, significantly enhancing adaptability across diverse driving conditions and vehicle types. We summarize our key contributions as:
\begin{itemize}
\item We propose an interpretable cooperative end-to-end driving framework by incorporating the \textit{Risk Map as Middleware} (RiskMM), which explicitly captures the spatiotemporal risk distribution. The risk map provides interpretable guidance for downstream planning.
\item The risk distribution is learned directly from multi-agent spatiotemporal scenario representations, supervised by planning trajectories. Then, an interpretable learning-based MPC planner is introduced to generate trajectories conditioned on the risk map under physical constraints.
\item We evaluate RiskMM on the largest real-world V2X sequential dataset, V2XPnP-Seq. Extensive experiments demonstrate its superior and robust performance on risk-aware planning tasks.
\end{itemize}

\section{Related Work}

\noindent \textbf{Interpretable End-to-end Autonomous Driving.} Despite the impressive driving performance demonstrated by end-to-end autonomous driving paradigms \cite{hu2023planning,liao2024diffusiondrive,zhou2025autovla}, their thorough end-to-end structure exacerbates the challenge of interpretability, especially for safety-critical driving systems that operate in the real physical world. Some approaches \cite{chen2021interpretable, chitta2021neat} incorporate auxiliary semantic tasks to enhance interpretability and provide additional supervision signals for end-to-end models. Other methods \cite{hu2022st, shao2023safety} employ multi-task learning to implicitly acquire a generalizable scenario representation, using auxiliary task outputs to jointly optimize planning and interpret the planning behavior. However, such interpretability remains limited to detection-level cues and lacks explicit modeling of interaction-induced risk and subsequent planning rationale. While large language models offer unified linguistic explanations for the end-to-end models \cite{xu2024drivegpt4, ding2024hint}, their inherent ambiguity prevents fine-grained, quantitative interpretation of behavior, such as that provided by risk maps or MPC weights. Moreover, language explanation introduces additional model complexity and demands high-quality reasoning data.

\noindent \textbf{Multi-agent End-to-end Autonomous Driving.} Due to occlusions and the limited perception range inherent in single-vehicle systems, multi-agent frameworks leveraging V2X communication have demonstrated promising potential \cite{zhou2024v2xpnp, zhou2022comprehensive}, though they remain in early development stages. V2X-Lead \cite{deng2023v2x} enhances the driving performance with deep reinforcement learning at unsignalized intersections with multi-agent LiDAR data. UniV2X \cite{yu2025end} exploits a cooperative end-to-end autonomous driving model with vehicle-to-infrastructure (V2I) to jointly improve perception, occupancy prediction, and planning, but it primarily operates on 2D visual inputs. More recently, V2X-VLM \cite{you2024v2x} incorporates V2I communication into an end-to-end driving pipeline via a vision-language model. However, despite these advances, the interpretability of cooperative end-to-end autonomous driving remains an open and underexplored challenge.


\noindent \textbf{Motion Planning.} 
Sampling-based\cite{xinyu2019bidirectional}, learning-based\cite{guo2024modeling}, and optimization-based \cite{hu2023lane, zhou2021reliable} planning approaches represent the three principal paradigms in motion planning. Sampling-based methods, while flexible, often neglect physical feasibility due to their non-parametric design and struggle to generate diverse driving behaviors. Learning-based methods address the latter limitation by capturing diverse behaviors from driving data but typically lack interpretability and physical constraints. However, optimization-based approaches explicitly model physical constrains, yet similarly fall short in capturing diverse behavior patterns. To this end, we introduce a learning-based MPC planner that integrates learnable cost functions and explicit vehicle dynamics.

\begin{figure*}[!t]
  \centering
  \includegraphics[width=0.9\linewidth]{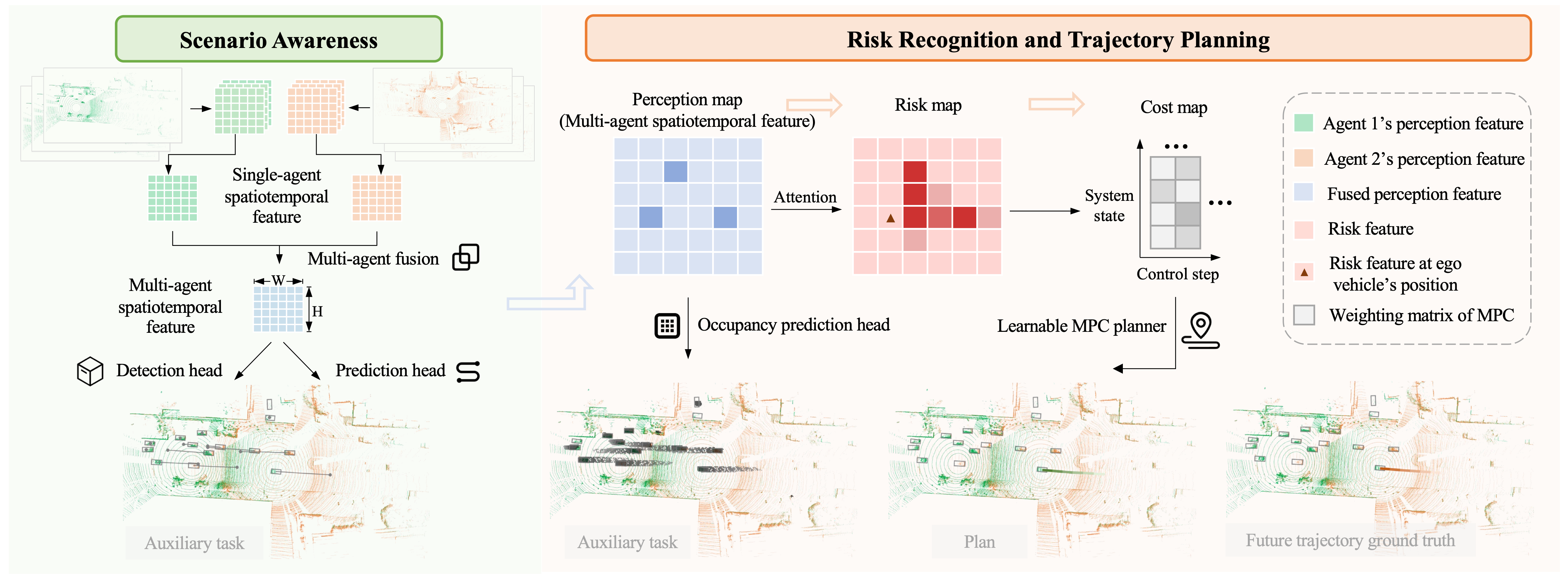}
  \caption{Overview of the proposed \textit{RiskMM} pipeline. \textit{(i) Scenario Awareness}: capturing a multi-agent spatiotemporal representation via auxiliary cooperative perception and prediction tasks; \textit{(ii) Risk Recognition and Trajectory Planning}: explicitly modeling the driving risk as a risk map and subsequently generating the planning trajectory with the learning-based MPC module.}
  \label{Figure2}
  \vspace{-0.6cm}
\end{figure*}

\section{METHODOLOGY}


To enhance interpretability in end-to-end autonomous driving, we introduce a logical reasoning framework that mirrors the cognitive process of human drivers, as illustrated in \cref{Figure2}, including scenario awareness and risk recognition.

\subsection{Scenario Awareness}

This module contributes to providing spatial-temporal features of the scenario to facilitate the downstream risk recognition task. It comprises seven stages: 

\noindent \textbf{V2X Metadata Sharing}. V2X metadata (such as pose and extrinsic parameters) are shared among all the agents (including CAVs and infrastructure units) to establish multi-agent spatial coordination in the scenario.

\noindent \textbf{LiDAR Feature Extraction}. Each agent extracts perception features from its own LiDAR data. Because of the low inference latency, PointPillar network \cite{lang2019pointpillars} is adopted to generate the BEV perception feature for each agent at each timestamp. The resulting LiDAR feature is structured as $\boldsymbol{F}_l\in \mathbb{R}^{T_{his} \times H \times W \times C}$ per agent, where $T_{his}$ represents the history horizon, $H$ and $W$ represent the number of height and width grid cells in the BEV representation respectively, and $C$ represents the feature dimension.

\noindent \textbf{Temporal Fusion}. To integrate historical information, the temporal sequence of LiDAR features from the same agent is fused along the time dimension, producing a consolidated feature $\boldsymbol{F}_t\in \mathbb{R}^{H \times W \times C}$ with a multi-head self-attention (MHSA) Transformer. To generate the time-aware representations for attention, LiDAR feature $\boldsymbol{F}_l$ is transformed as $\boldsymbol{F}_l^{'}=MLP(\boldsymbol{F}_{l}+\boldsymbol{P}_{l})$, where $\boldsymbol{P}_{l}$ denotes the embedding encoding historical timestamps. The fusion is formulated as:
\begin{equation}
    \boldsymbol{F}_{t}=M H S A\left(Q:\boldsymbol{F}_l^{'}, K:\boldsymbol{F}_l^{'}, V:\boldsymbol{F}_l^{'}\right).
    \label{Equation1}
\end{equation}

\noindent \textbf{Compression and Sharing}. Since each agent has generated its own temporal feature $\boldsymbol{F}_t$, inter-agent communication is required to aggregate global information. To reduce the transmission load, a convolution network is adopted to compress $\boldsymbol{F}_t$ before transmission, and then reconstructed by a convolution decoder at the receiving agent.

\noindent \textbf{Multi-agent Fusion}. Multi-agent fusion is required to achieve comprehensive scenario understanding. A Transformer-based method is adopted to realize it. $\boldsymbol{F}_i$ represents the $\boldsymbol{F}_t$ of agent $i$. The fused feature $\boldsymbol{F}_{s, i}$ for agent $i$ is formulated as:

$\boldsymbol{q}_{i}=M L P\left(\boldsymbol{F}_{i}\right), \boldsymbol{k}_{j}=M L P\left(\boldsymbol{F}_{j}\right), \boldsymbol{v}_{j}=M L P\left(\boldsymbol{F}_{j}\right),$
\begin{equation}
\begin{array}{c}
    \boldsymbol{F}_{s, i}=\sum\limits_{j} \operatorname{Softmax}\left(\boldsymbol{q}_{i} \cdot \boldsymbol{W}_{i j} \cdot \boldsymbol{k}_{j}\right) \cdot \boldsymbol{v}_{j},
    \end{array}
    \label{Equation2}
\end{equation}
where $j$ is the set of collaborating agents for agent $i$, and $\boldsymbol{W}_{i j}$ is the attention weight matrix encoding the spatial importance of agent $j$'s features to agent $i$.

\noindent \textbf{Map Feature Extraction}. The HD map information is necessary for prediction and planning. For each grid, its surrounding map polylines are extracted and accumulated, resulting in map feature $\boldsymbol{F}_m\in \mathbb{R}^{H \times W \times N_m \times N_w \times N_a}$. $N_m$ denotes the number of map polylines neighboring the given grid, $N_w$ denotes the number of waypoints for a map polyline, and $N_a$ denotes the number of attributes (such as position and lane type) for a waypoint. Max-pooling is adopted along the waypoint dimension to transform $\boldsymbol{F}_m$ into $\boldsymbol{F}_m^{'}\in \mathbb{R}^{H \times W \times N_m \times C}$. The map feature $\boldsymbol{F}_m^{'}$ is then concatenated with $\boldsymbol{F}_s\in \mathbb{R}^{H \times W \times 1 \times C}$, followed by an update of the BEV features using MHSA. By fusing LiDAR and HD map information, the final multi-agent spatiotemporal feature $\boldsymbol{F}_p\in \mathbb{R}^{C \times H \times W}$ is formulated.

\noindent \textbf{Auxiliary Cooperative Detection and Prediction Task}. The two auxiliary tasks are performed with two individual heads from multi-agent spatiotemporal feature $\boldsymbol{F}_p$. The feature is decoded into states for each predefined anchor box. For the detection head, two convolution layers are adopted to generate the regression state (including the position, size, and yaw of the bounding box) and the classification confidence score for each anchor box. The training loss includes $\mathcal{L}_1$ loss for regression and focal loss for classification. For the trajectory prediction head, two convolution layers are adopted to generate offset values for each anchor box at each timestamp. Then these offsets are accumulated to form a trajectory. The training loss is $\mathcal{L}_2$ loss between the predicted trajectories and the ground truth future trajectories.

\subsection{Risk Recognition}

Risk is defined as the potential spatial overlap between the ego agent and surroundings, which depends on two key factors: (i) the spatial layout of surroundings, provided by the perception module; and (ii) the likelihood of spatial conflict under interaction. To construct a fine-grained BEV risk map, we propose a two-stage risk quantification framework: \textit{(i) Occupancy Prediction}: For each BEV grid cell, occupancy features integrate the spatiotemporal feature and are extracted to encode the likelihood of being occupied by any traffic participant over time. \textit{(ii) Risk Map Generation}: To assess potential spatial conflicts between ego vehicle and surroundings, we model the interaction between the ego-occupied grid features and all other grid features. Specifically, attention mechanisms are employed to compute interaction scores, where higher attention weights indicate a greater likelihood of future collisions. As a result, the attention map serves as a quantitative risk map. Since the occupancy features integrate temporal features, the risk map is inherently predictive of future risks. The primary distinction between the occupancy map and the risk map is that the former is for perception, while the latter is for planning.

\noindent \textbf{Occupancy Prediction.} Multi-agent spatiotemporal feature $\boldsymbol{F}_p\in \mathbb{R}^{C \times H \times W}$ is respectively decoded into occupancy $\boldsymbol{O}\in \mathbb{R}^{T \times H \times W}$ and flow $\boldsymbol{f}\in \mathbb{R}^{(T-1) \times H \times W \times 2}$, where $T$ represents prediction horizon (including the current timestep), $2$ represents the number of grids moved in $H$ and $W$ direction. 
The flow prediction head adopts a Feature Pyramid Network (FPN) \cite{lin2017feature} to extract the movement features. The training loss for occupancy prediction includes binary cross-entropy and Dice loss, as the occupancy can be regarded as binary segmentation. The training loss for flow prediction is a weight-adapted loss, which is a variant of Smooth $\mathcal{L}_1$ Loss. When the ground truth of a specific grid is zero, the weight of Smooth $\mathcal{L}_1$ Loss is set to a low value $w_l$; otherwise, it is set to a high value $w_h$. In this paper, $w_l$ is 0.5 and $w_h$ is 2.0. This design adapts to the characteristic of occupancy flow ground truth, where zero values are frequently encountered.

\noindent \textbf{Risk Map Generation.} Multi-head cross-attention (MHCA) is adopted to capture the complex interaction. Each vehicle in the scenario takes turns serving as the ego vehicle and interacts with the entire scenario. Within the attention mechanism, the query is the feature of the grids that are currently occupied by vehicles, and the key is the feature of scenarios. The formulation is as follows:
\vspace{-0.4cm}
\begin{equation}
    \boldsymbol{F}_{r}, {W}_{r}=M H C A\left(Q:\left[\boldsymbol{F}_{v}, \boldsymbol{P}_{v}\right], K:\left[\boldsymbol{F}_{p}^{'}, \boldsymbol{P}_{p}\right], V: \boldsymbol{F}_{p}^{'}\right),
    \label{Equation3}
\end{equation}
where $\boldsymbol{F}_p^{'}\in \mathbb{R}^{(H \times W) \times 1 \times C}$ is the flattened representation of the multi-agent spatiotemporal feature $\boldsymbol{F}_p\in \mathbb{R}^{C \times H \times W}$. 
$\boldsymbol{P}_p\in \mathbb{R}^{(H \times W) \times 1 \times C}$ is the position embedding of $\boldsymbol{F}_p^{'}$, computed from the center points of the $H \times W$ spatial grids. $\boldsymbol{F}_{v}\in \mathbb{R}^{N \times 1 \times C}$ represents the subset of $\boldsymbol{F}_p^{'}$, consisting of grids currently occupied by vehicles. $N$ is the number of vehicles in the scenario. $\boldsymbol{P}_{v}\in \mathbb{R}^{N \times 1 \times C}$ is the position embedding of $\boldsymbol{F}_v$, computed from the center points of the $N$ spatial grids. $\boldsymbol{F}_{r}\in \mathbb{R}^{N \times 1 \times C}$ is the output of the attention layer, which represents the risk feature that facilitates risk-aware planning downstream. $\boldsymbol{W}_{r}\in \mathbb{R}^{1 \times N \times (H \times W)}$ is the attention weights of the attention layer, which represents the risk map. Along the $N$ dimension, each slice represents the risk distribution across the $H \times W$ grids for the ego perspective. The absence of ground-truth risk labels hinders conventional risk quantification. Our framework addresses this by learning risk end-to-end from comprehensive scenario representations, supervised by the planning trajectory.

\subsection{Trajectory Planning}


The key technical challenge lies in building a differentiable model to bridge the risk map and planning. To solve the problem, a planner's cost map is introduced. Through a perceptron layer, the risk feature $\boldsymbol{F}_{r}\in \mathbb{R}^{N \times 1 \times C}$ is decoded into the cost map $\boldsymbol{Z}\in \mathbb{R}^{N \times T \times M}$, which serves as the weighting parameters for the learning-based MPC planner, and $M$ represents the number of parameters in each time step. The learning-based MPC offers an interpretable planning process while facilitating the integration of physical constraints and explicit modeling of diverse vehicle dynamics. From an algebraic perspective, the nonlinear risk feature can be locally approximated by the quadratic objective function of the MPC planner. Specifically, in the vicinity of a nominal trajectory, a nonlinear function of system state perturbations can be approximated in a quadratic form.


\noindent \textbf{Risk-aware Planner.} Since the risk map provides temporal guidance over the prediction horizon, we adopt a learning-based MPC planner in our framework. For each specific vehicle, the planner's system state vector $\boldsymbol{X}$ is as follows:
\vspace{-0.3cm}
\begin{equation}
    \boldsymbol{X}_{k}=\left[\begin{array}{llll}
    s_{k} & v_{k} & l_{k} & \varphi_{k}
    \end{array}\right]^{T}, k \in[0, T-1],
    \label{Equation4}
\end{equation}
where $s_k$, $v_k$, $l_k$, and $\varphi_k$ represent the longitudinal position, speed, lateral position, and heading angle of the vehicle at control step $k$, which corresponds with the prediction step. $s_k$, $l_k$, and $\varphi_k$ are defined in the ego coordinate.

The planner's control vector $\boldsymbol{U}$ is as follows:
\vspace{-0.2cm}
\begin{equation}
    \boldsymbol{U}_{k}=\left[\begin{array}{ll}
    a_{k} & \delta_{k}
    \end{array}\right]^{T}, k \in[0, T-1],
    \label{Equation5}
\end{equation}
where $a_k$ and $\delta_{k}$ represent the acceleration and front wheel angle (ego coordinate) of the vehicle at control step $k$.

The planner's system dynamics is as follows:
\vspace{-0.2cm}
\begin{equation}
    \boldsymbol{X}_{k+1}=\boldsymbol{A}_{k} \boldsymbol{X}_{k}+\boldsymbol{B}_{k} \boldsymbol{U}_{k}, k \in[0, T-2],
    \label{Equation6}
\end{equation}
\begin{equation}
    \boldsymbol{A}_{k}=\Delta t \times\left[\begin{array}{cccc}
    0 & 1 & 0 & 0 \\
    0 & 0 & 0 & 0 \\
    0 & 0 & 0 & v_{k} \\
    0 & 0 & 0 & 0
    \end{array}\right]+\boldsymbol{I}_{4 \times 4}, k \in[0, T-2],
    \label{Equation7}
\end{equation}
\renewcommand{\arraystretch}{0.5}
\begin{equation}
    \boldsymbol{B}_{k}=\Delta t \times\left[\begin{array}{cc}
    0 & 0 \\
    1 & 0 \\
    0 & 0 \\
    0 & \frac{v_{k}}{l_{f r}}
    \end{array}\right], k \in[0, T-2],
    \label{Equation8}
\end{equation}
where $\Delta t$ denotes the time increment per control step, $\boldsymbol{I}$ is the identity matrix, and $l_{fr}$ represents the distance between the front and rear axles of the vehicle. As $l_{fr}$ varies across different vehicles, it is derived from the bounding box estimates provided by the scenario awareness module. This enables the planner to be vehicle-specific, enhancing the generalization capability. Moreover, this design leverages data from all vehicles in the scenario, effectively enlarging the training dataset beyond the ego CAV alone. 

The planner's cost function is as follows:
\begin{equation}
    J=\min \smash{\sum_{k=0}^{T-1}(\boldsymbol{X}_{k}^{T} \boldsymbol{Q}_{k}\boldsymbol{X}_{k}+\boldsymbol{U}_{k}^{T} \boldsymbol{R}_{k}\boldsymbol{U}_{k}+\boldsymbol{G}_{k}\boldsymbol{X}_{k}+\boldsymbol{H}_{k}\boldsymbol{U}_{k})},
    \label{Equation9}
\end{equation}
where $\boldsymbol{Q}_{k} \in \mathbb{R}^{4 \times 4}$, $\boldsymbol{R}_{k} \in \mathbb{R}^{2 \times 2}$, $\boldsymbol{G}_{k} \in \mathbb{R}^{1 \times 4}$ and $\boldsymbol{H}_{k} \in \mathbb{R}^{1 \times 2}$ ($k \in[0, T-1]$) are learnable weighting matrices. The parameters in these matrices are derived from the cost map $\boldsymbol{Z}\in \mathbb{R}^{N \times T \times M}$. In this paper, we set $M=14$. Specifically, for each vehicle at a given control step, the $M$ parameters are allocated as follows: 4 for the diagonal elements of $\boldsymbol{Q}_{k}$, 4 for $\boldsymbol{R}_{k}$, 4 for $\boldsymbol{G}_{k}$, and 2 for $\boldsymbol{H}_{k}$, where the learning results can be interpreted with the physical meaning of these weights in the MPC planner.

\begin{table*}[t]
\caption{Performance of \textit{RiskMM} with different settings in the real-world V2XPnP-Seq dataset.}
\vspace{-0.1cm}
\label{Table1}
\centering
\setlength{\tabcolsep}{3.2mm}
\renewcommand{\arraystretch}{1.05}
\begin{tabular}{ccc|cc|cc|cc} 
\toprule[1.1pt]
ID & Scenario awareness & Planning & AP@0.5 (\%)$\uparrow$ & EPA$\uparrow$ & AUC$\uparrow$ & Soft-IoU$\uparrow$ & ADE (m)$\downarrow$ & CR$\downarrow$ \\ 
\midrule 
0 & CooperRisk \cite{lei2025cooperrisk} & CooperRisk \cite{lei2025cooperrisk} & 58.0 & 0.400 & / & / & 3.345 & 0.149 \\
1 & FFNet \cite{yu2023flow} & Ours & 61.6 & 0.368 & 0.419 & 0.251 & 0.493 & 0.182 \\
2 & F-cooper \cite{chen2019f} & Ours & 68.1 & 0.435 & 0.507 & 0.341 & 0.434 & 0.163 \\
3 & V2X-ViT \cite{xu2022v2x} & Ours & 71.9 & 0.443 & 0.498 & 0.330 & 0.443 & 0.177 \\ 
\rowcolor{gray!20} \textbf{RiskMM*} & Ours & Safety-reinforced & 74.1& 0.423& 0.506& 0.341& 0.418 & \textbf{0.138}\\ 
\rowcolor{gray!20} \textbf{RiskMM} & Ours & Ours & \textbf{75.8} & \textbf{0.462} & \textbf{0.525} & \textbf{0.355} & \textbf{0.384}& 0.143 \\ 
\bottomrule[1.1pt]
\end{tabular}%
\vspace{-0.2cm}
\end{table*}

\begin{table*}[t]
\caption{Ablation results of \textit{RiskMM}.}
\vspace{-0.1cm}
\label{Table2}
\centering
\setlength{\tabcolsep}{4mm}
\renewcommand{\arraystretch}{1.05}
\begin{tabular}{cc|cc|cc|cc} 
\toprule[1.1pt]
Cooperative perception & Risk map & AP@0.5 (\%)$\uparrow$ & EPA$\uparrow$ & AUC$\uparrow$ & Soft-IoU$\uparrow$ & ADE (m)$\downarrow$ & CR$\downarrow$ \\ 
\midrule 
\ding{55} & \text{\checkmark} & 51.8 & 0.256 & 0.435 & 0.272 & 0.514 & 0.174 \\ 
\text{\checkmark} & \ding{55} & 75.8 & 0.462 & 0.525 & 0.355 & 2.868 & 0.340 \\ 
\rowcolor{gray!20} \text{\checkmark} & \text{\checkmark} & \textbf{75.8} & \textbf{0.462} & \textbf{0.525} & \textbf{0.355} & \textbf{0.384} & \textbf{0.143} \\ 
\bottomrule[1.1pt]
\end{tabular}%
\vspace{-0.2cm}
\end{table*}

\begin{table*}[t]
\caption{Performance of \textit{RiskMM} with different training strategies.}
\vspace{-0.1cm}
\label{Table3}
\centering
\setlength{\tabcolsep}{5mm}
\renewcommand{\arraystretch}{1.05}
\begin{tabular}{cc|cc|cc|cc} 
\toprule[1.1pt]
ID & Modular training & AP@0.5 (\%)$\uparrow$ & EPA$\uparrow$ & AUC$\uparrow$ & Soft-IoU$\uparrow$ & ADE (m)$\downarrow$ & CR$\downarrow$ \\ 
\midrule 
\multirow{2}{*}{\shortstack{1}} & \ding{55} & 57.1 & 0.339 & 0.415 & 0.240 & \textbf{0.469} & 0.213 \\
& \cellcolor{gray!20} \text{\checkmark} & \cellcolor{gray!20}\textbf{61.6} & \cellcolor{gray!20}\textbf{0.368} & \cellcolor{gray!20}\textbf{0.419} & \cellcolor{gray!20}\textbf{0.251} & \cellcolor{gray!20}0.493 & \cellcolor{gray!20}\textbf{0.182} \\
\midrule 
\multirow{2}{*}{\shortstack{2}} & \ding{55} & 62.4 & 0.416 & 0.403 & 0.258 & 0.936 & 0.369 \\
& \cellcolor{gray!20}\text{\checkmark} & \cellcolor{gray!20}\textbf{68.1} & \cellcolor{gray!20}\textbf{0.435} & \cellcolor{gray!20}\textbf{0.507} & \cellcolor{gray!20}\textbf{0.341} & \cellcolor{gray!20}\textbf{0.434} & \cellcolor{gray!20}\textbf{0.163} \\
\midrule 
\multirow{2}{*}{\shortstack{3}} & \ding{55} & 51.8 & 0.333 & 0.356 & 0.195 & 0.671 & 0.262 \\ 
& \cellcolor{gray!20}\text{\checkmark} & \cellcolor{gray!20}\textbf{71.9} & \cellcolor{gray!20}\textbf{0.443} & \cellcolor{gray!20}\textbf{0.498} & \cellcolor{gray!20}\textbf{0.330} & \cellcolor{gray!20}\textbf{0.443} & \cellcolor{gray!20}\textbf{0.177} \\ 
\midrule 
\multirow{2}{*}{\shortstack{\textbf{RiskMM}}} & \ding{55} & 63.6 & 0.423 & 0.298 & 0.178 & 0.394 & 0.161 \\ 
& \cellcolor{gray!20}\text{\checkmark} & \cellcolor{gray!20}\textbf{75.8} & \cellcolor{gray!20}\textbf{0.462} & \cellcolor{gray!20}\textbf{0.525} & \cellcolor{gray!20}\textbf{0.355} & \cellcolor{gray!20}\textbf{0.384} & \cellcolor{gray!20}\textbf{0.143} \\ 
\bottomrule[1.1pt]
\end{tabular}%
\vspace{-0.35cm}
\end{table*}

The proposed planner supports explicit incorporation of driving constraints, and we introduce a speed constraint:
\vspace{-0.4cm}
\begin{equation}
    |v_{k}|<v_{max}, k \in[0, T-1],
    \label{Equation10}
\end{equation}

where $v_{max}$ represents the maximum permitted vehicle speed. In this paper, we set $v_{max}=80mph$.
 
To ensure that the planner is differentiable, an explicit control policy is required. The Lagrange method is adopted to realize it, which reformulates the MPC model as a nonlinear optimization problem. The decision variable for the optimization problem is $\mathcal{\boldsymbol{X}}\in \mathbb{R}^{(4+2+4)T}$, which is the concatenation of the system state vector $\boldsymbol{X}$, the control vector $\boldsymbol{U}$ and the Lagrange Multiplier vector $\boldsymbol{\lambda}$ over the entire control horizon.
\vspace{-0.3cm}
\begin{equation}
    \mathcal{\boldsymbol{X}} = 
    \left[
        \underbrace{\boldsymbol{X}_0^{T}, \dots, \boldsymbol{X}_{T-1}^{T}}_{\text{states}},
        \underbrace{\boldsymbol{U}_0^{T}, \dots, \boldsymbol{U}_{T-1}^{T}}_{\text{controls}},
        \underbrace{\boldsymbol{\lambda}_0^{T}, \dots, \boldsymbol{\lambda}_{T-1}^{T}}_{\text{lagrange multipliers}}
    \right]^{T}.
    \label{Equation11}
\end{equation}

The control policy is computed as:
\begin{equation}
    \mathcal{\boldsymbol{X}}=-\left[\begin{array}{cc}
    \widetilde{\boldsymbol{Q}} & \widetilde{\boldsymbol{A}}^T \\
    \widetilde{\boldsymbol{A}} & \boldsymbol{0}
    \end{array}\right]^{-1}\left[\begin{array}{l}
    \widetilde{\boldsymbol{G}} \\
    \widetilde{\boldsymbol{B}}
    \end{array}\right],
    \label{Equation12}
\end{equation}
\begin{equation}
    \widetilde{\boldsymbol{Q}} = 
    \left[
    \begin{array}{ccc|ccc}
    \boldsymbol{Q}_0 & \mathbf{0} & \mathbf{0} & \mathbf{0} & \mathbf{0} & \mathbf{0} \\
    \mathbf{0} & \ddots & \mathbf{0} & \mathbf{0} & \mathbf{0} & \mathbf{0} \\
    \mathbf{0} & \mathbf{0} & \boldsymbol{Q}_{T-1} & \mathbf{0} & \mathbf{0} & \mathbf{0} \\
    \hline
    \mathbf{0} & \mathbf{0} & \mathbf{0} & \boldsymbol{R}_0 & \mathbf{0} & \mathbf{0} \\
    \mathbf{0} & \mathbf{0} & \mathbf{0} & \mathbf{0} & \ddots & \mathbf{0} \\
    \mathbf{0} & \mathbf{0} & \mathbf{0} & \mathbf{0} & \mathbf{0} & \boldsymbol{R}_{T-1}
    \end{array}
    \right],
    \label{Equation13}
\end{equation}
\begin{equation}
    \widetilde{\boldsymbol{A}} = 
    \left[
    \begin{array}{cccc|cccc}
    \mathbf{\boldsymbol{I}} & \mathbf{0} & \mathbf{0} & \mathbf{0} & \mathbf{0} & \mathbf{0} & \mathbf{0} & \mathbf{0} \\
    -\boldsymbol{A}_{0} & \mathbf{\boldsymbol{I}} & \mathbf{0} & \mathbf{0} & -\boldsymbol{B}_{0} & \mathbf{0} & \mathbf{0} & \mathbf{0} \\
    \mathbf{0} & \ddots & \mathbf{\boldsymbol{I}} & \mathbf{0} & \mathbf{0} & \ddots & \mathbf{0} & \mathbf{0} \\
    \mathbf{0} & \mathbf{0} & -\boldsymbol{A}_{T-2} & \mathbf{\boldsymbol{I}} & \mathbf{0} & \mathbf{0} & -\boldsymbol{B}_{T-2} & \mathbf{0}
    \end{array}
    \right],
    \label{Equation14}
\end{equation}
\begin{equation}
    \widetilde{\boldsymbol{G}} = 
    \left[
        -\boldsymbol{G}_0, \dots, -\boldsymbol{G}_{T-1},
        -\boldsymbol{H}_0, \dots, -\boldsymbol{H}_{T-1}
    \right]^{T},
    \label{Equation17}
\end{equation}
\begin{equation}
    \widetilde{\boldsymbol{B}} = 
    \left[
        -\boldsymbol{X}_0, \boldsymbol{0}
    \right].
    \label{Equation18}
\end{equation}


By executing the control policy under the imposed constraints, a planned trajectory is generated. This trajectory is derived from a subset of the system state vector $\boldsymbol{\mathcal{X}}$ and remains fully differentiable with respect to the end-to-end pipeline during training. Since $\widetilde{\boldsymbol{A}}$ is rank-deficient, a least-squares solution is used to solve Equation (\ref{Equation12}). By supervising the planned trajectories against ground-truth future trajectories, the model implicitly learns drivers’ subjective risk awareness, which is encoded within the risk map. The planning module is trained using Mean Squared Error (MSE) loss.

\begin{figure*}[t]
  \centering
  \includegraphics[width=\linewidth]{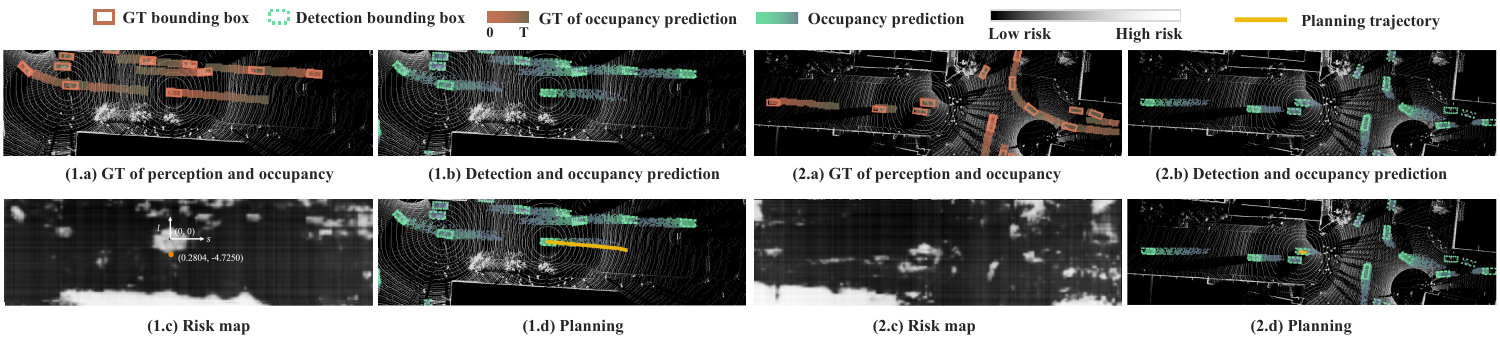}
  \caption{Qualitative results of \textit{RiskMM} in the real-world V2XPnP-Seq Dataset. The first and second rows show the performance in the first scenario, while the third and fourth rows illustrate the performance in the second scenario. The risk map aligns the information of the accurate detection and prediction, and introduces the interaction information to guide the planning.}
  \label{Figure3}
  \vspace{-0.4cm}
\end{figure*}

\begin{figure*}[t]
  \centering
  \includegraphics[width=\linewidth]{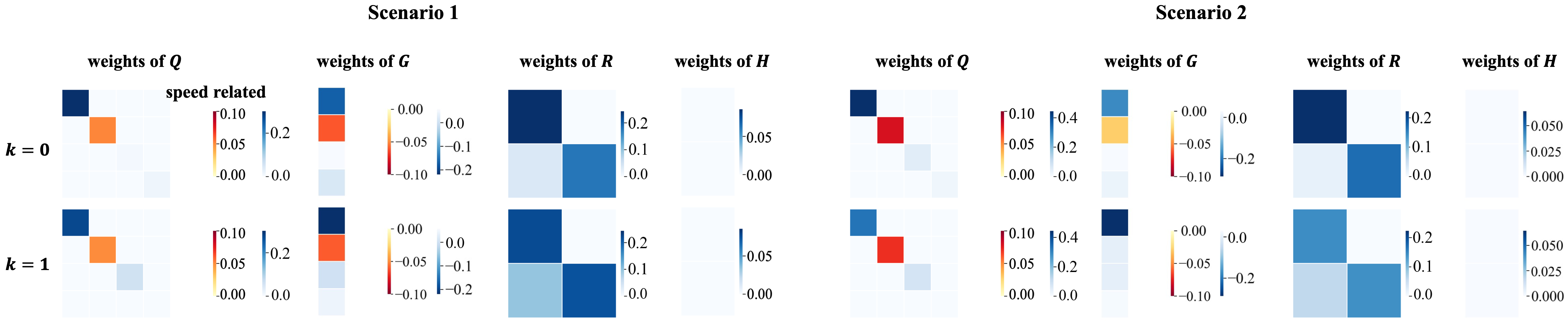}
  \caption{Learned weights of the proposed MPC planner. For Scenario 1, the quadratic term weight related to speed is relatively lower, while the linear term weight related to speed is relatively higher compared to Scenario 2.}
  \label{Figure4}
  \vspace{-0.8cm}
\end{figure*}

\section{EXPERIMENT}

\subsection{Evaluation Metrics}


To assess the performance of scenario awareness and risk-aware planning, we adopt a comprehensive set of metrics: Average Precision (AP) and End-to-end Perception and Prediction Accuracy (EPA) \cite{gu2023vip3d} for evaluating scene understanding; Area Under the Curve (AUC) and Soft Intersection-over-Union (Soft-IoU) for quantifying the occupancy prediction task; and Average Displacement Error (ADE) and Collision Rate (CR) for measuring planning accuracy and safety.

\noindent \textbf{Detection and Trajectory Prediction}. AP evaluates the detection performance with an intersection over the union threshold of 0.5. EPA evaluates the performance of joint perception and prediction:

\begin{equation}
    EPA=\smash{\frac{\left|\widehat{N}_{T P}\right|-\alpha N_{F P}}{N_{G T}}},
    \label{Equation19}
\end{equation}
where $|\widehat{N}_{T P}|$ is the number of true positive objects with prediction $minF D E<\tau_{E P A}$, $\tau_{E P A}$ is a threshold of prediction accuracy, $N_{F P}$, $N_{G T}$ are the number of false positive objects and ground truth objects, respectively, and $\alpha$ is a coefficient. $\tau_{E P A}$ is set to 2$\mathrm{m}$ and $\alpha$ is set to 0.5, following the settings in \cite{gu2023vip3d} and \cite{zhou2024v2xpnp}.

\noindent \textbf{Occupancy Prediction}. AUC is the area under the PR-curve. It is adopted because the occupancy of each grid cell could be treated as a separate binary prediction. Soft-IoU is the intersection-over-union between ground truth and predicted occupancy grids, which is calculated as:
\begin{equation}
    \text {Soft}\text{-}\mathrm{IoU}=\frac{\sum_{i=1}^{T} \frac{\sum_{j} O_{i}^{j} \widehat{O}_{i}^{j}}{\sum_{j} O_{i}^{j}+\widehat{O}_{i}^{j}-O_{i}^{j} \widehat{O}_{i}^{j}}}{T},
    \label{Equation20}
\end{equation}
where $O_{i}^{j}$ is the ground truth occupancy at prediction step $i$ for the $j^{th}$ grid cell, and $\widehat{O}_{i}^{j}$ is the predicted occupancy at prediction step $i$ for the $j^{th}$ grid cell.

\noindent \textbf{Planning}. 
ADE measures planning accuracy by computing the average L2 distance between the predicted trajectory of the ego vehicle and its ground-truth future trajectory over the prediction horizon. CR evaluates planning safety by calculating the proportion of scenarios where the distance between the ego vehicle's planned trajectory and other vehicles exceeds a predefined safety threshold. A scenario is considered to have exceeded the threshold if the time distance (Time-to-Collision, TTC) is less than $0.9\mathrm{s}$ \cite{dauner2024navsim} and the lateral distance is less than $3.5\mathrm{m}$.

\subsection{Baseline Methods}

\noindent \textbf{Generalization of \textit{RiskMM} to Multiple Tasks.} 
To evaluate the generalization capability of \textit{RiskMM}, we apply it across multiple tasks, including detection, occupancy prediction, and motion planning. Additionally, we extend it to a safety-reinforced variant, denoted as \textit{RiskMM*}, which emphasizes safety-aware planning rather than merely imitating driving behavior from the dataset. In this setting, the planner's loss function is augmented with a collision penalty term, while all other \textit{RiskMM} components remain unchanged.


\begin{figure*}[t]
  \centering
  \includegraphics[width=0.97\linewidth]{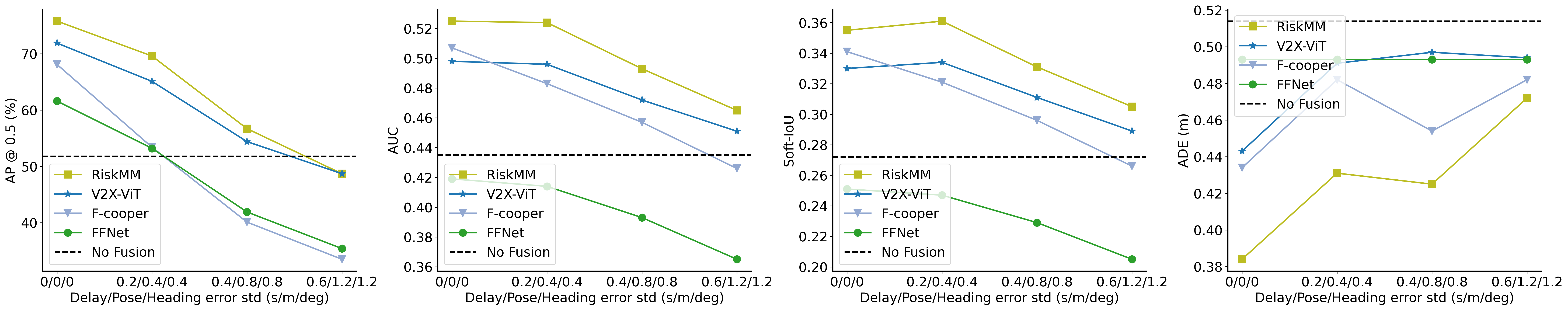}
  \caption{Communication noise and delay experiment.}
  \label{Figure5}
  \vspace{-0.8cm}
\end{figure*}

\noindent \textbf{Multi-agent Fusion.} 
To validate the effectiveness of our scenario awareness module, we integrate several multi-agent fusion method into the \textit{RiskMM} pipeline for comparison. Specifically, we adopt \textit{F-Cooper} \cite{chen2019f}, \textit{FFNet} \cite{yu2023flow}, and \textit{V2X-ViT} \cite{xu2022v2x} in \cref{Table1}, which represent state-of-the-art intermediate LiDAR fusion techniques. For fair comparison, all baselines are implemented using the same LiDAR backbone, decoding heads, and downstream modules as \textit{RiskMM}. We also introduce a no fusion baseline in \cref{Table2}.

\noindent \textbf{End-to-end Modelling.}
To validate the effectiveness of our proposed end-to-end design, a modular baseline is implemented. We adopt \textit{CooperRisk} \cite{lei2025cooperrisk} in \cref{Table1}. For a fair comparison, we replace its original, specially designed predictor with an LSTM network.

\noindent \textbf{Motion Planning Method.} To validate the effectiveness of our proposed risk map and learning-based MPC planner, a baseline without risk map is implemented. The motion planning in the baseline is replaced by a Convolutional Neural Network (CNN) based motion planning in \cref{Table2}.

\noindent \textbf{Training Strategy.} We introduce a \textit{RiskMM} baseline that directly trains the entire framework end-to-end, without modular decomposition for comparison in \cref{Table3}.

\subsection{Dataset and Implementation Details}

\noindent \textbf{Dataset.} V2XPnP-Seq dataset \cite{zhou2024v2xpnp} is utilized for the training and evaluation. It is the first real-world sequential V2X dataset, supporting all collaboration modes, collected by multiple mixed agents (i.e., two CAVs and two infrastructures). It covers 24 urban intersections, including roundabouts, T-junctions, and crossroads, totaling 40k frames of perception data and map data. The captured traffic is dense, with large-scale diverse interactive driving behaviors.

\noindent \textbf{Implementation Details.} Following the setting in \cite{zhou2024v2xpnp}, the evaluation range is $x \in[-70.4,70.4] \mathrm{m}$, $y \in[-40,40] \mathrm{m}$. The communication range is set to 50 $\mathrm{m}$. The historical horizon is 2 $\mathrm{s}$ (2 $\mathrm{Hz}$), and the future horizon for prediction and planning is 3 $\mathrm{s}$ (2 $\mathrm{Hz}$). Training proceeds in two stages: i) isolated training of the scenario awareness module, then ii) joint training of the entire pipeline. The optimizer uses AdamW with a weight decay of 0.01, and the learning rate scheduler adopts Cosine Annealing, initialized at $2\times e^{-4}$ with a minimum learning rate ratio of 0.05.

\subsection{Results Analysis}

The experimental results demonstrate that \textit{RiskMM} holds advantages across detection, prediction and planning compared to previous work CooperRisk \cite{lei2025cooperrisk}. These advantages validate the superiority of the end-to-end modelling design over the modular design. Table \ref{Table1} presents the performance of \textit{RiskMM} with different settings. Qualitative results are demonstrated in Fig. \ref{Figure3}.

\noindent \textbf{Interpretation.} The risk maps and their impacts on planning in two scenarios are illustrated in Fig. \ref{Figure3}. The results demonstrate that the risk map serves as an effective interpretation tool. It is capable of capturing the global traffic scenario and providing effective guidance for motion planning. The white regions in the illustrated risk maps indicate a high level of risk. Notably, the risk maps integrated temporal cues, which also serve as the attention of the planning task over the entire future horizon. In the first scenario, the risk map indicates potential conflicts with approaching vehicles from the front-left area. The ego vehicle pays heightened attention to the area and plans the trajectory to low-risk area. In the second scenario, the risk map identifies obstructed traffic participants ahead of the ego vehicle, causing the ego vehicle to stop.
To evaluate the interpretability of the proposed MPC planner, the weights in the planner's cost function are analyzed. For the first scenario, the absolute value of the linear term weight associated with vehicle speed is relatively higher and persists over more timesteps, indicating that the planner favors larger movements over a longer time horizon. In contrast, for the second scenario, the quadratic term weight related to speed is relatively higher, while the linear term weight related to speed is mostly zero, indicating that the planner tends to penalize movement. The learned weights for the two scenarios are illustrated in Fig. \ref{Figure4}.

\noindent \textbf{Generalization to Multiple Tasks.} 
The performance of \textit{RiskMM} across detection, occupancy prediction, and planning tasks are summarized in Table \ref{Table1}, which demonstrates its strong generalization and outperforms baseline methods in all tasks. Additionally, as compared to \textit{RiskMM}, \textit{RiskMM*} effectively enhances planning safety, reducing the Collision Rate (CR) from 0.143 to 0.138. Notably, this CR is lower than that of the dataset's ground-truth trajectories (CR = 0.142), indicating that \textit{RiskMM*} not only improves upon the baseline but can even surpass the human driver.


\noindent \textbf{Performance Enhancement through Multi-agent Fusion in Scenario Awareness.} As shown in \cref{Table1}, our scenario awareness method outperforms baseline methods. Compared to FFNet, our approach improves EPA, AUC, ADE of planning, and CR by 25.54\%, 25.30\%, 22.11\%, and 21.43\%, respectively. Compared to F-cooper, our approach improves EPA, AUC, ADE of planning, and CR by 6.21\%, 3.55\%, 11.52\%, and 12.27\%, respectively. Compared to V2X-ViT, our approach improves EPA, AUC, ADE of planning, and CR by 4.29\%, 5.42\%, 13.32\%, and 19.21\%, respectively. Additionally, as compared with single-agent setting in \cref{Table2}, \textit{RiskMM} achieves improvements of 46.33\% in AP and 80.47\% in EPA. Those results demonstrate that accurate scenario awareness from multiple agents contributes to the performance of downstream tasks.


\noindent \textbf{Ablation Study on Risk Map.} \cref{Table2} shows that incorporating the risk map enhances planning performance in both ADE and CR. It can even help single-agent systems outperform multi-agent systems without the risk map module. The risk map augments the motion planner with explicit risk context and provides interpretable results.

\noindent \textbf{Ablation Study on Modular Training}. \cref{Table3} demonstrates that the proposed pipeline with modular training improves the performance in diverse tasks. It attains a 19.18\% increase in AP, 9.22\% increase in EPA, 76.17\% increase in AUC, 99.44\% increase in Soft-IoU, 2.54\% decrease in ADE of planning, and 11.18\% decrease in CR. Although end-to-end learning performs better on individual tasks for some baselines, the superiority is unstable.

\noindent \textbf{Robustness Assessment on V2X Noise Settings}. In order to evaluate the robustness of \textit{RiskMM}, particularly the degradation under varying noise levels, we injected artificial noise and latency. Sensitivity analysis on detection, occupancy prediction and planning under different communication noise is presented in Figure \ref{Figure5}. The positional and heading noises are drawn from a Gaussian distribution with standard deviations ranging from $0.2\mathrm{m}$ to $1.0\mathrm{m}$ for positional noise and from $0.2^{\circ}$ to $1.0^{\circ}$ for heading noise, respectively\cite{xu2022v2x, zhou2024v2xpnp}. The time delay is configured to range from $100\mathrm{ms}$ to $500\mathrm{ms}$. It is demonstrated that \textit{RiskMM} demonstrates superior robustness, maintaining optimal performance across all noise conditions.

\section{CONCLUSIONS}

In this paper, we proposed \textit{RiskMM}, an interpretable cooperative end-to-end cooperative driving framework leveraging a \textit{Risk Map as Middleware}. By explicitly modeling the spatiotemporal risk distribution derived from the multi-agent spatiotemporal representation of the scenario, RiskMM enhances both the interpretability and safety of the downstream planning module. Moreover, a interpretable learning-based MPC module integrates explicit physical constraints and vehicle parameters, facilitating transparency and quality of planning. Extensive experiments on the V2XPnP-Seq dataset demonstrated RiskMM's effectiveness, surpassing baseline methods in multiple tasks and maintaining strong robustness and interpretability.
\textit{RiskMM} represents a step toward more interpretable and adaptable cooperative automation.

{
\renewcommand{\baselinestretch}{0.95}
\normalsize
\bibliographystyle{IEEEtran}
\bibliography{root}
}

\end{document}